\documentclass{article}

    \usepackage[preprint]{neurips_2023}

\usepackage[utf8]{inputenc} % allow utf-8 input
\usepackage[T1]{fontenc}    % use 8-bit T1 fonts
\usepackage{hyperref}       % hyperlinks
\usepackage{url}            % simple URL typesetting
\usepackage{amsmath}
\usepackage{booktabs}       % professional-quality tables
\usepackage{amsfonts}       % blackboard math symbols
\usepackage{nicefrac}       % compact symbols for 1/2, etc.
\usepackage{microtype}      % microtypography
\usepackage{xcolor}         % colors
\usepackage {graphicx}
\usepackage{multirow}
\usepackage{array}
\usepackage{booktabs}
\usepackage{amssymb}
\usepackage{bbding}
\usepackage{pifont}
\usepackage{wasysym}
\usepackage{utfsym}
\usepackage{fontawesome}
\usepackage{geometry}
\usepackage{rotating}
\usepackage{amsmath}
\usepackage{amsfonts}
\usepackage{fancyvrb}
\usepackage[ruled,vlined]{algorithm2e}
\usepackage{algorithmic}
\usepackage{amsmath}
\usepackage{colortbl} % \cellcolor
\usepackage{caption}
\usepackage{wrapfig}
\usepackage{pifont}
\usepackage{subfiles}
\usepackage{enumitem}
\usepackage{bm}
\usepackage{footmisc}
\setlist[itemize]{itemsep=2pt,parsep=0pt,topsep=2pt,partopsep=0pt}
\newcommand{\ie}{\textit{i}.\textit{e}.}
\newcommand{\eg}{\textit{e}.\textit{g}.}
\newcommand{\cf}{\textit{cf.}}

\definecolor{Cerulean}{RGB}{0,123,167}
\hypersetup{colorlinks,linkcolor={blue},citecolor={green},urlcolor={magenta}}

\definecolor{mypink}{rgb}{1.0, 0.6, 0.8}
\definecolor{mygreen}{rgb}{0.0, 0.7, 0.0}
\definecolor{myblue}{rgb}{0.0, 0.72, 0.86}

\setcitestyle{square,comma,numbers}

\title{Zero-shot Visual Relation Detection via Composite Visual Cues from Large Language Models}

\author{%
  Lin Li$^{1,2}$, Jun Xiao$^1$, Guikun Chen$^1$, Jian Shao$^1$, Yueting Zhuang$^1$, Long Chen$^2$\thanks{Long Chen is the corresponding author. Work was done when Lin Li visited HKUST.} \\
  \small $^1$Zhejiang University \; 
  $^2$The Hong Kong University of Science and Technology \\
  \small \texttt{\{mukti, junx, guikun.chen, jshao, yzhuang\}@zju.edu.cn, longchen@ust.hk} \\
  \href{https://github.com/HKUST-LongGroup/RECODE}{https://github.com/HKUST-LongGroup/RECODE} \\
  }
\begin{document}

\maketitle

\begin{center}
    \centering
    \includegraphics[width=1.0\linewidth]{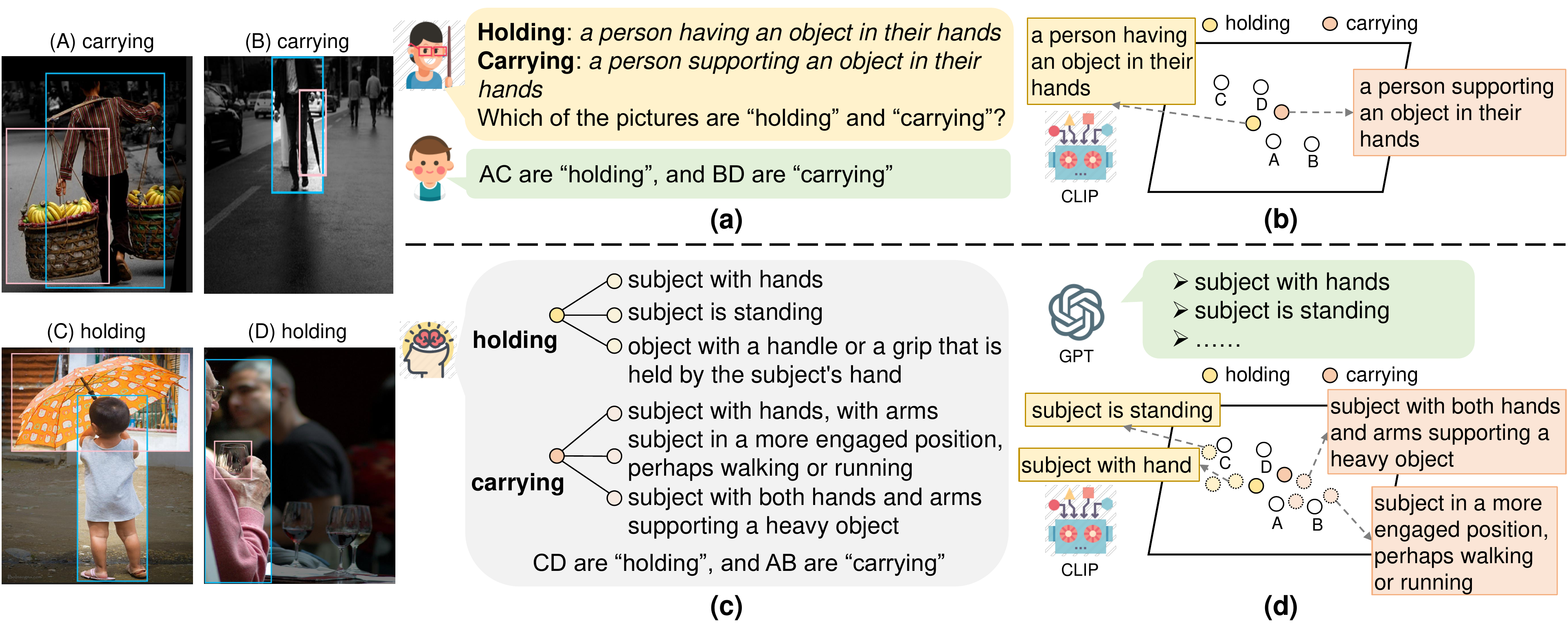}
    \captionof{figure}{Illustration of the challenges of VRD with similar relation categories \textbf{holding} and \textbf{carrying}. Four images and their ground-truths are on the left. The subject and object for each triplet are denoted by \textcolor{myblue}{\textbf{blue}} and \textcolor{mypink}{\textbf{pink}} boxes, respectively. \textbf{(a)} A child may incorrectly identify these two relations only based on similar concepts alone. \textbf{(b)} Using class-based prompts, CLIP always maps these two relations to adjacent locations in the semantic space. \textbf{(c)} We humans always utilize composite visual cues to correctly distinguish between different relations. \textbf{(d)} Our proposed RECODE uses LLM (\eg, GPT) to generate composite descriptions that aid the CLIP model in distinguishing between them.}
    \label{fig:1}
\end{center}%

\begin{abstract}

Pretrained vision-language models, such as CLIP, have demonstrated strong generalization capabilities, making them promising tools in the realm of zero-shot visual recognition. Visual relation detection (VRD) is a typical task that identifies relationship (or interaction) types between object pairs within an image. However, naively utilizing CLIP with prevalent \emph{class-based} prompts for zero-shot VRD has several weaknesses, \eg, it struggles to distinguish between different fine-grained relation types and it neglects essential spatial information of two objects. To this end, we propose a novel method for zero-shot VRD: \textbf{RECODE}, which solves RElation detection via COmposite DEscription prompts. Specifically, RECODE first decomposes each predicate category into subject, object, and spatial components. Then, it leverages large language models (LLMs) to generate description-based prompts (or visual cues) for each component. Different visual cues enhance the discriminability of similar relation categories from different perspectives, which significantly boosts performance in VRD. To dynamically fuse different cues, we further introduce a chain-of-thought method that prompts LLMs to generate reasonable weights for different visual cues. Extensive experiments on four VRD benchmarks have demonstrated the effectiveness and interpretability of RECODE.
\end{abstract}

\section{Introduction}
\label{sec:intro}

Recent advances in pretrained vision-language models (VLMs)~\citep{radford2021learning,wortsman2022robust,Cho2022CLIPReward,jia2021scaling} (\eg, CLIP~\citep{radford2021learning}), have shown remarkable generalization ability and achieved impressive performance on zero-shot recognition tasks. Specifically, CLIP employs two encoders: an image encoder that converts images into visual features, and a text encoder that transforms sentences into semantic features. This design allows the encoders to map different modalities into a common semantic space. When the inputs to the text encoder are \emph{class-based prompts}, such as ``A [CLASS]'', ``A photo of [CLASS]'', CLIP can compare the image and prompts in the shared semantic space, thereby enabling zero-shot recognition of novel categories~\citep{radford2021learning}.
Compared to object recognition, visual relation detection (VRD) is much more challenging, which needs to identify the relation types between object pairs within an image in the form of $\langle$\texttt{subject}, \texttt{relation}, \texttt{object}$\rangle$~\citep{xu2017scene,tang2020unbiased,li2022rethinking,hetao2022,li2023compositional}. It differs from object recognition in that it requires an understanding of how objects are related to each other. By crafting class-based prompts to describe these relation types, CLIP could potentially be extended to perform zero-shot VRD.

\begin{figure*}[!t]
  \centering
    \includegraphics[width=1\linewidth]{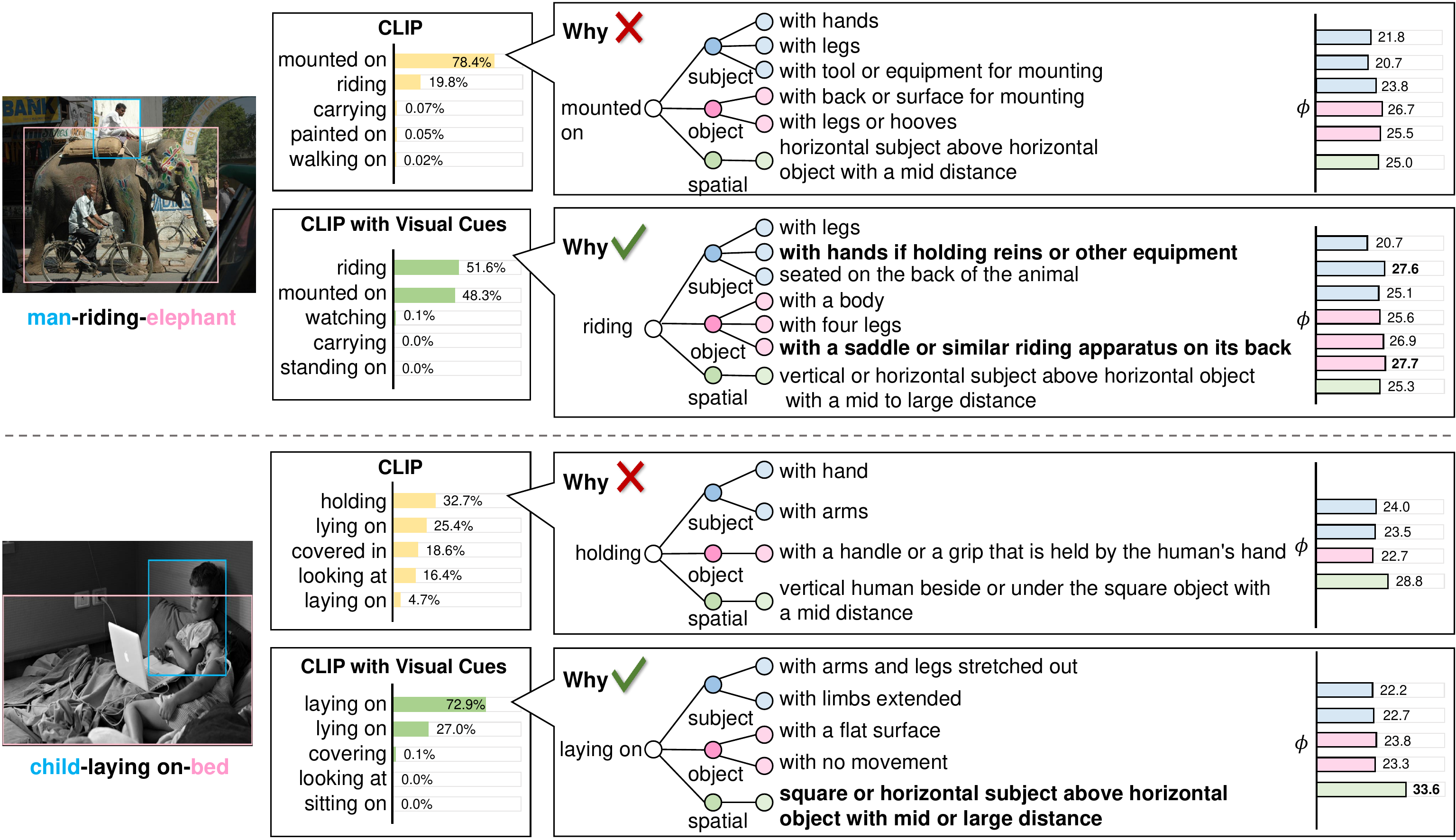}
    \vspace{-1.5em}
    \caption{A comparative analysis of predictions made by RECODE and baseline CLIP using class-based prompts. It illustrates how our method offers interpretability to the relation classification results through the similarity $\phi$ between the image and the description-based prompts.}
    \label{fig:2}
    \vspace{-1em}
\end{figure*}

However, this straightforward baseline presents notable challenges. Imagine you are a child asked to distinguish relation categories ``holding'' and ``carrying'', both involving a person and an object. Based on the similar concepts of ``holding'' (\ie, a person having an object in their hands) and ``carrying'' (\ie, a person supporting an object in their hands), it would be difficult to determine the correct prediction (\cf, Figure~\ref{fig:1}(a)). In other words, class-based prompts of ``holding'' and ``carrying'' might be projected to adjacent locations in semantic space by CLIP, leading to a \textbf{relation sensitivity} issue: CLIP struggles to differentiate between the subtle nuances of similar relations.
Secondly, class-based prompts overlook the unique spatial cues inherent to each relation category, leading to a \textbf{spatial discriminability} issue. The ``holding'' category generally suggests the object being at a certain height and orientation relative to the person, while ``carrying'' implies a different spatial position, typically with the object located lower and possibly supported by the person's entire body. The neglect of spatial cues leads to inaccuracies in distinguishing between such spatial-aware relation categories. Moreover, applying CLIP in this manner brings about a \textbf{computational efficiency} issue. Using CLIP requires cropping each union region of a subject-object pair separately from the original image (\ie, $N^2$ crops for $N$ proposals), leading to computational inefficiencies.

Nonetheless, we humans can distinguish relation categories from different visual cues. For example, from the subject's perspective, we could think that in the case of ``holding'', a person might be standing while having an object, such as an umbrella, in their hand. Meanwhile, in the case of ``carrying'', a person should be in a more engaged position, perhaps walking or running with both hands and arms supporting a heavy object, like a suitcase. In addition, spatial cues also play an important role in identifying these relation categories. For example, when a person is carrying an umbrella, the umbrella is usually positioned lower and closer to the person's body compared to when the person is holding an umbrella. Based on these visual cues, we can easily identify scenarios such as ``\texttt{person-holding-umbrella}'' and ``\texttt{person-carrying-umbrella}'' as in Figure~\ref{fig:1}(c).

Inspired by our humans' ability to extract and utilize different visual cues, we present a novel method for zero-shot VRD: \textbf{RECODE}, which classifies RElation via COmposite DEscriptions. It first uses large language models (LLMs)~\citep{brown2020language}, to generate detailed and informative descriptions\footnote{We use a description to represent a visual cue.\label{footnote:description}} for different components of relation categories, such as subject, object, and spatial. These descriptions are then used as description-based prompts for the CLIP model, enabling it to focus on specific visual features that help distinguish between similar relation categories and improve VRD performance. Specifically, 
for the subject and object components, these prompts include visual cues such as appearance (\eg, with leg), size (\eg, small), and posture (\eg, in a sitting posture). For the spatial component, these prompts include cues related to the spatial relationships between objects, such as relative position and distance. By incorporating different visual cues, RECODE enhances the discriminability of similar relation categories, such as ``riding'' and ``mounted'' based on the different postures of the subject, \eg, ``seated on the back of animal'' for the subject of ``riding''. Similarly, spatial visual cues can be used to differentiate between ``laying on'' and ``holding'' based on the relative position between the subject and object, such as ``subject above object'' and ``subject under object'' (\cf, Figure~\ref{fig:2}).

In addition, we explore the limitations of several description generation prompts for visual cue, \eg, relation class description prompt~\citep{menon2022visual}, and then design a guided relation component description prompt that utilizes the high-level object categories to generate more accurate visual cues for each relation category. For instance, if the high-level category of object is ``animal'', the generated object descriptions for relation ``riding'' are tailored to the ``animal'' category, \eg, ``with four legs'', instead of the ``product'', \eg, ``with wheels''. Meanwhile, to better fuse the evidence from different visual cues, we further leverage LLMs to predict reasonable weights for different components. Particularly, we design a chain-of-thought (CoT) method~\citep{kojima2022large} to break down this weight assignment problem into smaller, more manageable pieces, and prompt LLM to generate a series of rationales and weights.

To evaluate our RECODE, we conducted experiments on four benchmark datasets: Visual Genome (VG)~\citep{krishna2017visual} and GQA~\citep{hudson2019gqa} datasets for scene graph generation (SGG), and HICO-DET~\citep{chao2015hico} and V-COCO~\citep{gupta2015visual} datasets for human-object interaction (HOI) detection. Experimental results prove the generalization and interpretability of our method. In summary, we made three main \textbf{contributions} in this paper: 1) We analyze the weaknesses of the prevalent class-based prompt for zero-shot VRD in detail and propose a novel solution RECODE. RECODE leverages the power of LLMs to generate description-based prompts (visual cues) for each component of the relation class, enhancing the CLIP model's ability to distinguish between various relation categories. 2) We introduce a chain-of-thought method that breaks down the problem into smaller, more manageable pieces, allowing the LLM to generate a series of rationales for each cue, ultimately leading to reasonable weights for each component. 3) We conduct experiments on four benchmark datasets and demonstrate the effectiveness and interpretability of our method.

\section{Approach}
Typically, VRD is comprised of two sub-tasks: object detection and relation classification~\citep{xu2017scene}. Since zero-shot object detection has been extensively studied~\citep{yan2022semantics,radford2021learning, menon2022visual}, in this paper, we primarily focus on zero-shot relation classification. Specifically, given the bounding boxes (bboxes) $\{b_i\}$ and object categories $\{o_i\}$ of all objects, our target is to predict the visual relation (or predicate/interaction) categories $\{r_{ij}\}$ between pairwise objects. To facilitate presentation, we use $s$, $o$, and $p$ to denote the subject, object, and their spatial position in a triplet respectively, and $r$ to denote the relation category. 

\noindent{\textbf{Class-based Prompt Baseline for Zero-Shot VRD.}} Following recent zero-shot object recognition methods, a straightforward solution for zero-shot VRD is the CLIP with class-based prompt. Specifically, a pretrained CLIP image encoder $V(\cdot)$ and a pretrained CLIP text encoder $T(\cdot)$ are used to classify pairwise objects with a set of relation classes. For each relation class, a natural language \textbf{class-based prompt} $p_c$ is generated, incorporating the relation information, \eg, ``[REL-CLS]-ing/ed'' or ``a photo of [REL-CLS]''. Each prompt is then passed through $T(\cdot)$ to get semantic embedding $t$, while the union region of a subject-object pair is passed through $V(\cdot)$ to get visual embedding $v$. The cosine similarity between $v$ and $t$ of different relation categories is calculated and processed by a softmax function to obtain the probability distribution over all relation categories.

\begin{figure*}[!t]
  \centering
  \includegraphics[width=1.0\linewidth]{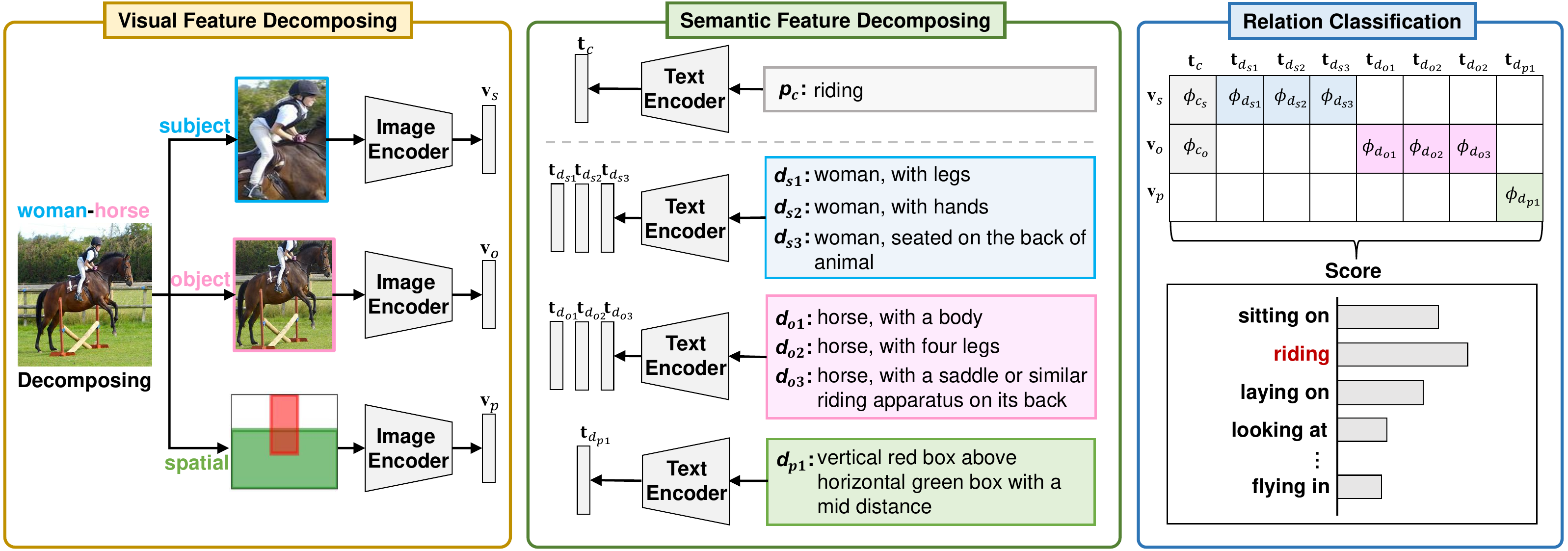} 
  \vspace{-1.5em}
  \caption{The framework of RECODE. 1) \textbf{\emph{Visual feature decomposing}} decomposes the triplet into subject, object, and spatial features. 2) \textbf{\emph{Semantic feature decomposing}} decomposes relation categories into subject, object, and spatial descriptions. 3) \textbf{\emph{Relation classification}} calculates similarities between decomposed visual and semantic features and applies softmax to obtain the probability distribution.}
  \vspace{-1em}
  \label{fig:framework}
\end{figure*}

\subsection{Zero-shot VRD with Composed Visual Cues}
\label{approach}

To overcome the limitations of class-based prompts, we propose a novel approach RECODE for zero-shot VRD. It consists of three parts: visual feature decomposing, semantic feature decomposing, and relation classification (\cf, Figure~\ref{fig:framework}). In the first two parts, we decompose the visual features of the triplet into subject, object, and spatial features, and then generate semantic features for each component. In the last part, we calculate the similarities between the decomposed visual features and a set of semantic features, and aggregate them to get the final predictions over all relations.

\noindent{\textbf{Visual Feature Decomposing.}
To enhance spatial discriminability and computational efficiency, we decompose the visual features of a triplet into subject, object, and spatial features. \textit{For subject and object features}, we crop the regions of the subject and object from the original image using the given bboxes $b_s$ and $b_o$, and encode them into visual embeddings $v_s$ and $v_o$ using the image encoder $V(\cdot)$ of CLIP. \textit{For spatial features}, we aim to obtain the spatial relationship between the subject and object based on their bounding boxes. However, directly obtaining all spatial images based on the given bounding boxes is computationally expensive due to the diversity of spatial positions ($N^2$ each image). To address this, we simulate the spatial relationship between the subject and object using a finite set of spatial images, represented by red and green bboxes respectively. We define four attributes (shape, size, relative position, and distance) based on bounding box properties. Each attribute is assigned a finite set of values to construct a finite set of simulated spatial images. For a given triplet, we match the calculated attribute values with the most similar simulated image\footnote{Due to the limited space, the details are left in the appendix. \label{footnote:appendix}}. The matched spatial image is then encoded into a visual embedding $v_p$ using $V(\cdot)$ of CLIP.

\noindent{\textbf{Semantic Feature Decomposing.}} To improve the CLIP model's ability to distinguish between different relation classes, we incorporate a set of \textbf{description-based prompts} $D$ to augment the original class-based prompt for each relation category. \textit{For the subject and object components}, we generate a set of description-based prompts $D_s$ and $D_o$ to provide additional visual cue information, the generation process is described in Sec.~\ref{generate}. These prompts contain object categories with specific visual cues that highlight the unique characteristics of the relation being performed, \eg, ``women, with legs'', which enhances the discriminability between similar relation categories. \textit{For the spatial component}, it only contains a set of description-based prompts $D_p$ that include information about the relative position and distance between the subject and object in the image. By incorporating this additional information, we aim to distinguish between relations based on spatial location. After generating these sets of description-based prompts, we obtain semantic embeddings $\{t_{{d_{{s_i}}}}\}$, $\{t_{{d_{{o_i}}}}\}$, and $\{t_{{d_{{p_i}}}}\}$ using a text encoder $T(\cdot)$, separately. These embeddings, along with the class-based prompt embedding $t_c$, are used for relation classification.

\begin{figure*}[!t]
  \centering
  \includegraphics[width=1.0\textwidth]{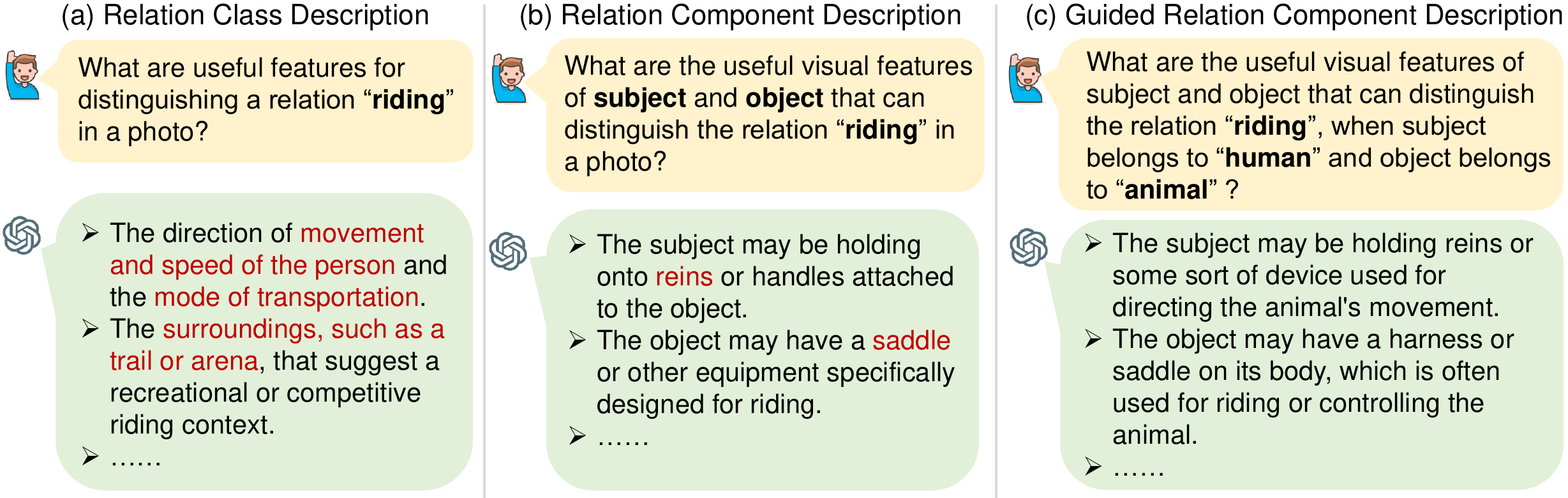}
  \vspace{-1.5em}
  \caption{Examples of different prompts used for generating descriptions of visual cues. (a) \textbf{\emph{Relation class description}} generates descriptions for each relation class directly. (b) \textbf{\emph{Relation component description}} generates descriptions for each component of the relation separately. (c) \textbf{\emph{Guided relation component description}} incorporates high-level object category guide generation process.}
  \label{fig:prompt}
  \vspace{-0.5em}
\end{figure*}

\noindent{\textbf{Relation Classification.}}
In this step, we compute the similarity score between the visual and semantic features to obtain the relation probability distribution. We first calculate the cosine similarity $\phi(\cdot,\cdot)$ between each visual embedding and semantic embedding for each relation category $r$. The final score incorporates both class-based and description-based prompts, and is calculated as follows:
\begin{equation}
\begin{small}
{S}(r) = \underbrace{\phi({v_s},{t_c}) + \phi({v_o},{t_c})}_{\text{class-based prompts}} + \underbrace{\sum\limits_{k
\in\{s,o,p\}} {\frac{{w_k}}{{|{D_k}(r)|}} \left[ {\textstyle{\sum\limits_{{d_{k_{i}}}\in{D_{k}(r)}}} {\phi({v_k},{t_{{d_{k_{i}}}}})}}\right]}}_{\text{description-based prompts}} ,
\end{small}
\end{equation}
where $w_k$ represents the importance of visual cues for each component $k\in\{s,o,p\}$, and $|{D_k}(r)|$ denotes the number of visual cues in ${D_k}(r)$ for relation category $r$. We compute the similarity of individual visual cues for each component and then obtain their average. The weights of different components are determined by a LLM, which will be discussed in Sec.~\ref{generate}. Finally, we apply a softmax operation to the scores to obtain the probability distribution over all relation categories.

\subsection{Visual Cue Descriptions and Weights Generation}
\label{generate}

LLMs, such as GPT~\citep{brown2020language}, have been shown to contain significant world knowledge. In this section, we present the process of generating descriptions of visual cue $D_s$, $D_o$, and $D_p$, as well as the weights $w_s$, $w_o$, and $w_p$ for each component of each relation category using LLMs. 

\subsubsection{Visual Cue Descriptions} 
In this section, we explore methods for generating descriptions of visual cues for relation decomposition. Inspired by the work~\cite{menon2022visual} of zero-shot image classification, we first propose \textbf{relation class description} prompt, which generates descriptions from the perspective of class-level (\cf, Figure~\ref{fig:prompt}(a)). It has the advantage of producing descriptions that are easy to interpret and understand. However, it may result in overly diverse and information-rich descriptions that could hinder the extraction of meaningful visual cues, \eg, ``speed of the person'' in Figure~\ref{fig:prompt}(a).

To address this limitation, we then consider another \textbf{relation component description} prompt, which involves decomposing the relation into its subject and object components and generating descriptions of their visual features separately (\cf, Figure~\ref{fig:prompt}(b)). While this type of prompt allows for more focused and specific descriptions of visual cues, it may not be effective in capturing the variations in visual features between different subject-object category pairs. For example, ``\texttt{man-riding-horse}'' and ``\texttt{person-riding-bike}'' typically have totally different visual features for the object. The visual cues ``reins'' and ``saddle'' of the object in Figure~\ref{fig:prompt}(b) are inappropriate for a ``bike''.

Therefore, we design \textbf{guided relation component description} prompt. It builds upon the second method by incorporating the high-level category information of the object into the generation process, leading to more accurate and informative descriptions of the visual features of both the subject and object components (\cf, Figure~\ref{fig:prompt}(c)). To achieve this, we classify the object into high-level classes, such as ``human'', ``animal'', and ``product'', to guide the description generation. For example, ``bike'' is classified as ``product'', and ``horse'' is classified as ``animal''. This allows for the separate generation of visual feature descriptions for each high-level object class, \eg, ``a harness or saddle on its body'' for ``animal'', resulting in more precise and relevant visual cues for each relation category\footref{footnote:appendix}.

\begin{figure*}[!t]
  \centering
\includegraphics[width=1.0\textwidth]{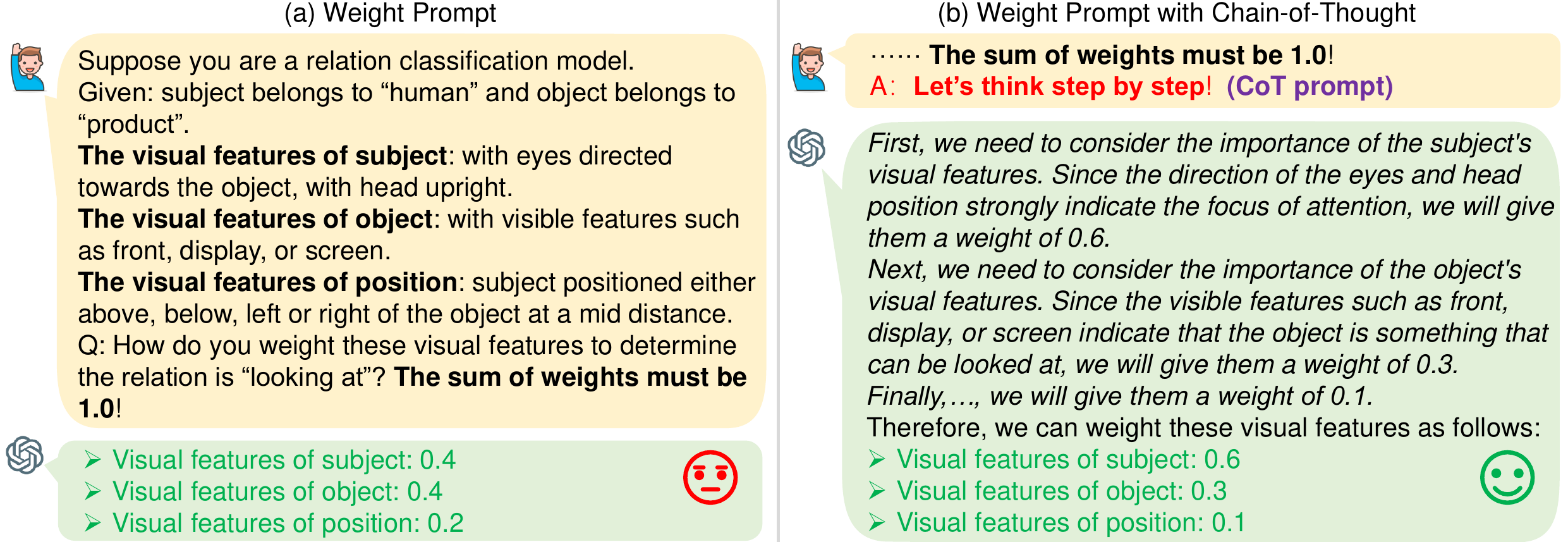}
\vspace{-1.5em}
\caption{Illustration of the effectiveness of CoT method in generating reasonable visual cue weights. (a) \textbf{Prompt without CoT}. LLM assigns same weights for subject and object. (b) \textbf{Prompt with CoT}. LLM analyzes the importance of each cue step by step and assigns more reasonable weights.}
\vspace{-0.5em}
  \label{fig:cot}
\end{figure*}

\subsubsection{Visual Cue Weights}
Intuitively, different combinations of visual cues may have varying degrees of importance in relation classification. For example, for relation ``looking at'', the visual cue ``with visible features'' of the object may not be as informative as the visual cue ``with eye'' of the subject. To account for this, we leverage the impressive knowledge and reasoning abilities of LLMs to analyze the discriminative power of different visual cues and dynamically assign weights accordingly. Specifically, we provide each combination of visual cues as input to LLM and prompt it to determine the appropriate weight for each cue for distinguishing the given predicate. The prompts used for this purpose are in Figure~\ref{fig:cot}.

\noindent\textbf{Chain-of-Thought (CoT) Prompting.}
To ensure the generated weights are reasonable, we utilize a CoT method that has demonstrated remarkable reasoning abilities~\citep{kojima2022large,zhang2022automatic}. Specifically, we prompt the LLM to generate rationales by using the stepwise reasoning prompt ``\textbf{Let's think step by step!}'' to break down the problem into smaller, more manageable pieces. Then LLM generates a series of rationales, and those that lead to the reasonable weights. For example in Figure~\ref{fig:cot}, we demonstrate the importance of the CoT method in generating more accurate weights. Without the stepwise reasoning prompt, LLM generates the same weight for both the subject and object visual cues for ``looking at'', which is clearly unreasonable. However, with the CoT prompt, LLM is able to analyze each cue step by step, leading to a more accurate assignment of weights, \ie, the cues about the subject are relatively more important. In order to standardize the format of the strings generated by LLMs for extracting different components of visual cues and weights, we make certain modifications to the prompts for descriptions and weights\footref{footnote:appendix}.

\section{Experiment}
\subsection{Experiment setup}

\noindent\textbf{Datasets.} We evaluated our method on four zero-shot VRD benchmarks: 1) \textbf{VG}~\citep{krishna2017visual} contains 26,443 images for testing, each annotated with object and predicate labels to form a scene graph. Following previous works~\citep{xu2017scene}, we used the pre-processed VG with 150 object classes. We adopted the 24 semantic predicate classes proposed in~\citep{zellers2018neural,bui2022sg}, as they are more informative and challenging for classifying.
2) \textbf{GQA}~\citep{hudson2019gqa} is a large-scale SGG dataset. We used the same split provided by~\citep{dong2022stacked}, which contains 8,208 images for testing with 200 object classes. As for predicate classes, we selected 26 semantic predicate classes by referring to VG.
3) \textbf{HICO-DET}~\citep{chao2015hico} contains 9,658 testing images annotated with 600 HOI triplets derived from combinations of 117 verb classes and 80 object classes. 
4) \textbf{V-COCO}~\citep{gupta2015visual} comprises 4,946 testing images annotated with 29 action categories.

\noindent\textbf{Evaluation Metrics.} For SGG datasets (\ie, VG and GQA), we reported \textbf{Recall@K (R@K)} which indicates the proportion of ground-truths that appear among the top-K confident predictions, and \textbf{mean Recall@K (mR@K)} which averages R@K scores calculated for each category separately~\cite{li2022devil}. For HOI datasets (\ie, HOI-DET and V-COCO), we reported \textbf{mean Average Precision (mAP)}~\cite{chao2018learning}.

 \addtolength{\tabcolsep}{-1.5pt}
\begin{table}[!t]
  \centering
  \renewcommand\arraystretch{1.0}
  \setlength\tabcolsep{2pt}
  \caption{Evaluation results on the test set of VG and GQA datasets. 
  % CLS, CLSDE, and RECODE denote using prompts of relation CLasS, relation CLasS DEscription, and guided RElation COmponent DEscription, respectively. 
  $\dag$ denotes removing the guidance from high-level object category. $\star$ denotes integrated with Filter strategy.}
    \begin{tabular}{c|l|cc|cc|cc|cc|cc|cc}
    \specialrule{0.1em}{0pt}{0pt}
    \hline
    \multicolumn{1}{c|}{\multirow{2}[4]{*}{Data}} & \multicolumn{1}{c|}{\multirow{2}[4]{*}{Method}} & \multicolumn{12}{c}{Predicate Classification} \\
\cline{3-14}    \multicolumn{1}{c|}{} &       & \small{R@20}  & \textcolor[rgb]{ .753,  0,  0}{$\scriptstyle \bigtriangleup$} & \small{R@50}  & \textcolor[rgb]{ .753,  0,  0}{$\scriptstyle \bigtriangleup$} & \small{R@100} & \textcolor[rgb]{ .753,  0,  0}{$\scriptstyle \bigtriangleup$} & \small{mR@20} & \textcolor[rgb]{ .753,  0,  0}{$\scriptstyle \bigtriangleup$} & \small{mR@50} & \textcolor[rgb]{ .753,  0,  0}{$\scriptstyle \bigtriangleup$} & \small{mR@100} & \textcolor[rgb]{ .753,  0,  0}{$\scriptstyle \bigtriangleup$} \\
    \hline
    \multirow{5}[2]{*}{\begin{sideways}VG\end{sideways}} & CLS  & 7.2   & \textcolor[rgb]{ .753,  0,  0}{-}     & 10.9  & \textcolor[rgb]{ .753,  0,  0}{-}     & 13.2  & \textcolor[rgb]{ .753,  0,  0}{-}     & 9.4   & \textcolor[rgb]{ .753,  0,  0}{-}     & 14.0    & \textcolor[rgb]{ .753,  0,  0}{-}     & 17.6  & \textcolor[rgb]{ .753,  0,  0}{-} \\
          & CLSDE & 7.0   & \textcolor[rgb]{ .753,  0,  0}{-0.2 } & 10.6  & \textcolor[rgb]{ .753,  0,  0}{-0.3 } & 12.9  & \textcolor[rgb]{ .753,  0,  0}{-0.3 } & 8.5   & \textcolor[rgb]{ .753,  0,  0}{-0.9 } & 13.6  & \textcolor[rgb]{ .753,  0,  0}{-0.4 }  & 16.9  & \textcolor[rgb]{ .753,  0,  0}{-0.7 } \\
          & RECODE$^\dagger$ & 7.3   & \textcolor[rgb]{ .753,  0,  0}{0.1} & 11.2  & \textcolor[rgb]{ .753,  0,  0}{0.3} & 15.4  & \textcolor[rgb]{ .753,  0,  0}{2.2} & 8.2   & \textcolor[rgb]{ .753,  0,  0}{-1.2 } & 13.5  & \textcolor[rgb]{ .753,  0,  0}{-0.5 }  & 18.3  & \textcolor[rgb]{ .753,  0,  0}{0.7} \\
          & RECODE & \textbf{9.7} & \textcolor[rgb]{ .753,  0,  0}{2.5} & \textbf{14.9} & \textcolor[rgb]{ .753,  0,  0}{4.0} & \textbf{19.3} & \textcolor[rgb]{ .753,  0,  0}{6.1} & \textbf{10.2} & \textcolor[rgb]{ .753,  0,  0}{0.8} & \textbf{16.4} & \textcolor[rgb]{ .753,  0,  0}{2.4} & \textbf{22.7} & \textcolor[rgb]{ .753,  0,  0}{5.1} \\
          & RECODE$^\star$ & \textbf{10.6} & \textcolor[rgb]{ .753,  0,  0}{3.4} & \textbf{18.3} & \textcolor[rgb]{ .753,  0,  0}{7.4} & \textbf{25.0} & \textcolor[rgb]{ .753,  0,  0}{11.8} & \textbf{10.7} & \textcolor[rgb]{ .753,  0,  0}{1.3} & \textbf{18.7} & \textcolor[rgb]{ .753,  0,  0}{4.7} & \textbf{27.8} & \textcolor[rgb]{ .753,  0,  0}{10.2} \\
    \hline
    \multirow{5}[2]{*}{\begin{sideways}GQA\end{sideways}} & CLS  & 5.6   & \textcolor[rgb]{ .753,  0,  0}{-}     & 7.7   & \textcolor[rgb]{ .753,  0,  0}{-}     & 9.9   & \textcolor[rgb]{ .753,  0,  0}{-}     & 6.3   & \textcolor[rgb]{ .753,  0,  0}{-}     & 9.5   & \textcolor[rgb]{ .753,  0,  0}{-}    & 12.2  & \textcolor[rgb]{ .753,  0,  0}{-} \\
          & CLSDE & 5.4   & \textcolor[rgb]{ .753,  0,  0}{-0.2 } & 7.2   & \textcolor[rgb]{ .753,  0,  0}{-0.5 } & 9.3   & \textcolor[rgb]{ .753,  0,  0}{-0.6 } & 6.0   & \textcolor[rgb]{ .753,  0,  0}{-0.3 } & 8.8   & \textcolor[rgb]{ .753,  0,  0}{-0.7 }  & 11.5  & \textcolor[rgb]{ .753,  0,  0}{-0.7 } \\
          & RECODE$^\dagger$ & 5.2   & \textcolor[rgb]{ .753,  0,  0}{-0.4 } & 7.8   & \textcolor[rgb]{ .753,  0,  0}{0.1} & 10.2  & \textcolor[rgb]{ .753,  0,  0}{0.3} & 5.8   & \textcolor[rgb]{ .753,  0,  0}{-0.5 } & 8.9   & \textcolor[rgb]{ .753,  0,  0}{-0.6 }  & 11.3  & \textcolor[rgb]{ .753,  0,  0}{-0.9 } \\
          & RECODE & \textbf{6.3} & \textcolor[rgb]{ .753,  0,  0}{0.7} & \textbf{9.4}   & \textcolor[rgb]{ .753,  0,  0}{1.7} & \textbf{11.8} & \textcolor[rgb]{ .753,  0,  0}{1.9} & \textbf{7.8} & \textcolor[rgb]{ .753,  0,  0}{1.5} & \textbf{11.9} & \textcolor[rgb]{ .753,  0,  0}{2.4}   & \textbf{15.1} & \textcolor[rgb]{ .753,  0,  0}{2.9} \\
          & RECODE$^\star$ & \textbf{7.0} & \textcolor[rgb]{ .753,  0,  0}{1.4} & \textbf{11.1} & \textcolor[rgb]{ .753,  0,  0}{3.4} & \textbf{15.4} & \textcolor[rgb]{ .753,  0,  0}{5.5} & \textbf{9.4} & \textcolor[rgb]{ .753,  0,  0}{3.1} & \textbf{14.8} & \textcolor[rgb]{ .753,  0,  0}{5.3} & \textbf{20.4} & \textcolor[rgb]{ .753,  0,  0}{8.2} \\
    \specialrule{0.1em}{0pt}{0pt}
    \hline
    \end{tabular}%
  \label{tab:sgg_res}%
  \vspace{-1.0em}
\end{table}%
 \addtolength{\tabcolsep}{1.5pt}

 % \addtolength{\tabcolsep}{-1.5pt}
 % \begin{table}[!t]
 %   \centering
 %   \renewcommand\arraystretch{1.1}
 %  \setlength\tabcolsep{3pt}
 %   \caption{Evaluation results on the test set of HICO-DET and V-COCO dataset.}
 %     \begin{tabular}{l|ccc|cc}
 %     \specialrule{0.1em}{0pt}{0pt}
 %     \hline
 %    \multicolumn{1}{c|}{\multirow{2}[4]{*}{Method}} & \multicolumn{3}{c|}{HICO-DET} & \multicolumn{2}{c}{V-COCO} \\
 %    \cline{2-6}    \multicolumn{1}{c|}{} & \multicolumn{1}{l}{Full} & \multicolumn{1}{l}{Rare} & \multicolumn{1}{l|}{Non-Rare} & \multicolumn{1}{c}{Role 1} & \multicolumn{1}{c}{Role 2} \\
 %    \hline
 %    CLS  & 32.3  & 33.2  & 31.8  & 25.5  & 28.6 \\
 %    CLSDE & 32.5  & 33.1  & 32.2  & 25.6  & 28.8 \\
 %    RECODE$\dagger$ & 32.5  & 33.0    & 32.4  & 25.7  & 28.8 \\ 
 %    RECODE & \textbf{32.7} & \textbf{33.2} & \textbf{32.5} & \textbf{26.0} & \textbf{29.0} \\
 %     \specialrule{0.1em}{0pt}{0pt}
 %     \hline
 %     \end{tabular}%
 %   \label{tab:hoi_res}%
 % \end{table}%
 %  \addtolength{\tabcolsep}{1.5pt}

\noindent\textbf{Implementation Details.} For the LLM, we employed the GPT-3.5-turbo, a highly performant variant of the GPT model. As for CLIP, we leveraged the OpenAI's publicly accessible resources, specifically opting for the Vision Transformer with a base configuration (ViT-B/32) as default backbone\footref{footnote:appendix}.

% \noindent\textbf{Implementation Details.} We employed GPT-3.5-turbo for the LLM and used the ViT-B/32 configuration as the default backbone for the CLIP model\footref{footnote:appendix}.

\noindent\textbf{Settings.} The bounding box and category of objects were given in all experiments. We compared our RECODE with two baselines: 1) \textbf{CLS}, which uses relation-CLasS-based prompts (\eg, ``riding'') to compute the similarity between the image and text. 2) \textbf{CLSDE}, which uses prompts of relation CLasS DEscription as shown in Figure~\ref{fig:prompt}(a). Each component of the proposed framework can serve as a plug-and-play module for zero-shot VRD. Specifically, 1) \textbf{Filter}, which denotes filtering those unreasonable predictions (\eg, \texttt{kid-eating-house}) with the rules generated by GPT\footref{footnote:appendix}. 2) \textbf{Cue}, which denotes using description-based prompts (Sec.~\ref{approach}). 3) \textbf{Spatial}, which denotes using spacial images as additional features. 4) \textbf{Weight}, which denotes using dynamic weights generated by GPT to determine the importance of each feature, \ie, visual cue weights.

\subsection{Results and Analysis}
In this work, we evaluated the prediction performance of the proposed framework on two related tasks, \ie, SGG and HOI. The former outputs a list of relation triplet \texttt{$\langle$sub,pred,obj$\rangle$}, while the latter just fix the category of \texttt{sub} to human. Overall, our method
achieved significant improvement on the two tasks compared to the CLS baseline, which shows the superiority of our method.

 \begin{wraptable}{r}{6.5cm}
   \centering
   \renewcommand\arraystretch{1.00}
  \setlength\tabcolsep{2pt}
  \vspace{-1em}
   \caption{Evaluation results on the test set of HICO-DET and V-COCO datasets.}
     \vspace{-0.7em}
        \scalebox{0.94}{
     \begin{tabular}{l|ccc|cc}
     \specialrule{0.1em}{0pt}{0pt}
     \hline
    \multicolumn{1}{c|}{\multirow{2}[4]{*}{Method}} & \multicolumn{3}{c|}{HICO-DET} & \multicolumn{2}{c}{V-COCO} \\
    \cline{2-6}    \multicolumn{1}{c|}{} & \small{Full} & \small{Rare} & \small{Non-Rare} & \small{Role 1} & \small{Role 2} \\
    \hline
    CLS  & 32.3  & 33.2  & 31.8  & 25.5  & 28.6 \\
    CLSDE & 32.5  & 33.1  & 32.2  & 25.6  & 28.8 \\
    RECODE$^\dagger$ & 32.5  & 33.0    & 32.4  & 25.7  & 28.8 \\ 
    RECODE & \textbf{32.7} & \textbf{33.2} & \textbf{32.5} & \textbf{26.0} & \textbf{29.0} \\
     \specialrule{0.1em}{0pt}{0pt}
     \hline
     \end{tabular}%
  }
    \vspace{-0.6em}
   \label{tab:hoi_res}%
 \end{wraptable}
 
\noindent\textbf{Evaluation on SGG.} 
From the results in Table~\ref{tab:sgg_res}, we have the following observations: 1) CLSDE showed worse performance than the trivial CLS baseline. This is because the considerable noise in CLSDE which may hinder the model to attend the most distinguishable parts. 2) With the proper guidance, RECODE achieved considerable improvements compared to the baselines, \eg, 0.8\% to 6.1\% gains on VG and 0.7\% to 2.9\% gains on GQA. The performance drops of RECODE$\dag$ also demonstrated the importance of guidance from high-level object categories during the generation process. 3) Integrated with the filtering strategy, \ie, RECODE$^\star$, achieved the best performance over all metrics, which suggests that commonsense knowledge is complementary and effective for zero-shot VRD. It also demonstrated that CLIP struggles to distinguish abstract concepts, \ie, relation sensitivity as mentioned in Sec.~\ref{sec:intro}. 

\noindent\textbf{Evaluation on HOI.} 
Since standard evaluation procedure of HOI had filtered out those unreasonable predictions, RECODE$^\star$ was not evaluated here. From the results in Table~\ref{tab:hoi_res}, we can observe that the performance gains were lower than those on SGG, \eg, 0.0\% to 0.7\% gains on HICO-DET and 0.4\% to 0.5\% gains on V-COCO. The reasons are two-fold. On the one hand, since the category of subject is always a human, its features are too similar to be distinguished by CLIP. On the other hand, some of the actions are very similar in appearance. For example, distinguishing between actions like ``\texttt{person-throw-sports ball}'' and ``\texttt{person-catch-sports ball}'' is challenging due to their visual similarity.

\addtolength{\tabcolsep}{-1.5pt}
\begin{table}[!t]
  \centering
  \caption{Ablation studies on different architectures of CLIP. The official released weights are used.}
    \renewcommand\arraystretch{0.95}
 \setlength\tabcolsep{3pt} 
    \begin{tabular}{l|l|ccc|ccc}
    \specialrule{0.1em}{0pt}{0pt}
    \hline
    \multicolumn{1}{l|}{\multirow{2}[4]{*}{Architecture}} & \multirow{2}[4]{*}{Method} & \multicolumn{6}{c}{Predicate Classification} \\
\cline{3-8}          &       & \small{R@20}  & \small{R@50}  & \small{R@100} & \small{mR@20} & \small{mR@50} & \small{mR@100} \\
    \hline
    \multicolumn{1}{l|}{\multirow{2}[2]{*}{ViT-L/14}} &  CLS$^\star$ & 8.3   & 15.0    & 21.5  & 7.6   & 14.2  & 24.2 \\
    \multicolumn{1}{l|}{} & RECODE$^\star$ & \textbf{11.2} & \textbf{19.9} & \textbf{28.0} & \textbf{9.1} & \textbf{18.5} & \textbf{28.1} \\
    \hline
    \multirow{2}[2]{*}{ViT-L/14@336px} & CLS$^\star$ & 8.6   & 15.4  & 21.8  & 7.7   & 13.9  & 23.0 \\
          & RECODE$^\star$ & \textbf{12.1} & \textbf{21.1} & \textbf{29.2} & \textbf{9.7} & \textbf{19.5} & \textbf{28.2} \\
    \hline    
    \multirow{2}[2]{*}{ViT-B/32} & CLS$^\star$ & 7.5   & 13.7  & 19.4  & 9.1   & 15.9  & 24.0 \\
          & RECODE$^\star$ & \textbf{10.6} & \textbf{18.3} & \textbf{25.0} & \textbf{10.7} & \textbf{18.7} & \textbf{27.8} \\

    \hline
    \multirow{2}[2]{*}{ViT-B/16} & CLS$^\star$ & 8.6   & 15.5  & 22.1  & 9.8   & 17.2  & 25.2 \\
          & RECODE$^\star$ & 
    \textbf{12.6} & \textbf{21.0} & \textbf{28.5} & \textbf{12.5} & \textbf{20.2} & \textbf{30.0} \\
    \specialrule{0.1em}{0pt}{0pt}
    \hline
    \end{tabular}%
  \label{tab:ab_archi}%
  \vspace{-1.0em}
\end{table}%
\addtolength{\tabcolsep}{1.5pt}

% \addtolength{\tabcolsep}{-1.5pt}

% \addtolength{\tabcolsep}{1.5pt}

\subsection{Diagnostic Experiment}

\noindent\textbf{Architectures.}
We investigated the impact of changing the architectures of CLIP, as shown in Table~\ref{tab:ab_archi}. From the results, we can observe consistent improvements regardless of the architecture used. 

\begin{wraptable}{r}{9.2cm}
   \centering
   \renewcommand\arraystretch{1.0}
  \setlength\tabcolsep{0.8pt}
  \vspace{-1em}
   \caption{Analysis of key components on the test set of VG. %Filter, which filters out commonsense-inconsistent \texttt{sub-pred} and \texttt{pred-obj} categories using GPT; Spatial, which incorporates spatial visual and semantic features; and Weight, which uses the CoT method to generate weights for the visual cues of each component.
  }
     \vspace{-0.5em}
     \scalebox{0.94}{
     \begin{tabular}{c|c|c|c|ccc|ccc}
     \specialrule{0.1em}{0pt}{0pt}
     \hline
     \multirow{2}[3]{*}{\small{Filter}} & \multirow{2}[3]{*}{\small{Cue}} & \multirow{2}[3]{*}{\small{Spatial}} & \multirow{2}[3]{*}{\small{Weight}} & \multicolumn{6}{c}{Predicate Classification} \\
     \cline{5-10}  &       &    &   & \small{R@20} & \small{R@50} & \small{R@100} & \small{mR@20} & \small{mR@50} & \small{mR@100} \\
     \hline
           &  &     &     & 7.2   & 10.9  & 13.2  & 9.4   & 14.0    & 17.6 \\
           & \ding{51}   &    &       & 7.4   & 12.3  & 16.6  & 9.0   & 14.0  & 19.5  \\
           &   \ding{51} &\ding{51}  &       & 9.1   & 13.4  & 17.4  & 9.3   & 15.0  & 20.3  \\
           &  \ding{51} &     &  \ding{51} & 7.9   & 13.4  & 17.7  & 9.3   & 14.7  & 20.5  \\
           & \ding{51} & \ding{51}  &  \ding{51}  & \textbf{9.7} & \textbf{14.9} & \textbf{19.3} & \textbf{10.2} & \textbf{16.4} & \textbf{22.7} \\
     \hline
      \ding{51}  &  &    &      & 7.5   & 13.7  & 19.4  & 9.1   & 15.9  & 24.0 \\
     \ding{51} & \ding{51} &       &       & 8.8   & 15.9  & 23.5  & 10.3  & 17.2  & 26.2  \\
     \ding{51} & \ding{51} &  \ding{51}  &       & 9.3   & 16.3  & 22.5  & 10.1  & 18.1  & 25.5  \\
     \ding{51} & \ding{51} &       &  \ding{51}  & 10.0  & 17.5  & 24.8  & 10.4  & 17.8  & 26.7  \\
    \ding{51} &\ding{51} &  \ding{51}  &  \ding{51}  & \textbf{10.6} & \textbf{18.3} & \textbf{25.0} & \textbf{10.7} & \textbf{18.7} & \textbf{27.8} \\
      \hline
     \specialrule{0.1em}{0pt}{0pt}
     \hline
     \end{tabular}%
 }
       \vspace{-0.5em}
   \label{tab:sgg_ab}%
\end{wraptable}
\noindent\textbf{Key Component Analysis.} The results are summarized in Table~\ref{tab:sgg_ab}. The first row refers to the CLS baseline. Four crucial conclusions can be drawn. \textbf{First}, with the guidance of Cue, consistent improvements can be observed, \eg, 0.2\% to 3.4\% gains on R@K w/o Filter and 1.3\% to 4.1\% gains on R@K with Filter. \textbf{Second}, by introducing the spatial feature, the relative position of subject and object is considered, resulting in notable performance gains on R@K (0.8\% to 1.7\%) and mR@K (0.3\% to 1.0\%) w/o Filter compared to just using Cue. This is because the spatial feature is of importance for relation detection~\cite{tang2020unbiased}. \textbf{Third}, benefiting from the impressive reasoning ability of LLMs, the proposed weighting strategy can determine the importance of different cues, thus achieving further improvements, \eg, 0.5\% to 1.1\% gains on R@K compared to average aggregation. \textbf{Fourth}, by filtering those unreasonable predictions, consistent improvements can be observed. The reason may be that the performance of relation detection of CLIP is not accurate enough. Empirically, commonsense knowledge is a feasible way to filter those noise. Combining all components allows for the best overall performance on all evaluation metrics.

\noindent\textbf{Case study.} To investigate the most important regions for distinguishing relations, we visualized the attention map given different images and prompts (\cf, Figure~\ref{fig:vis}). From the visualization of class-based prompts, we can observe that CLIP may attend those regions unrelated to the query prompts, \eg, focusing on the body of a person given relation ``growing on''. We attribute this phenomenon to the insufficient information within given prompts, which is also our motivation to introduce visual cue descriptions. As for description-based prompts, CLIP can attend to right regions with the guidance of descriptions, \eg, focusing on colorful patterns on the product given relation ``painted on''.

\begin{figure*}[!t]
  \centering
      \includegraphics[width=1.0\linewidth]{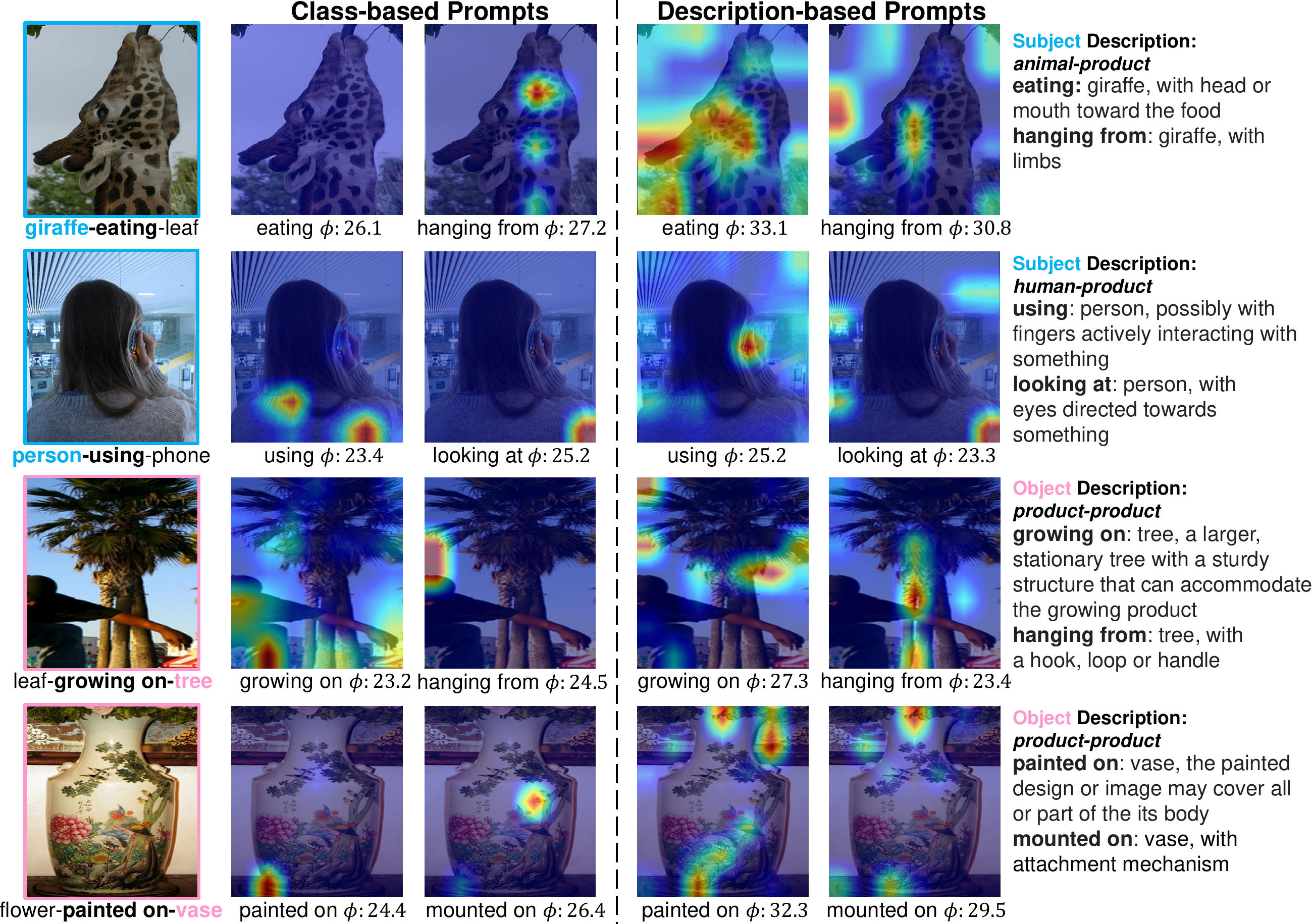} 
  \vspace{-1.5em}
  \caption{Visualization of CLIP attention maps on input images with different prompts. The right side shows the partial description-prompts generated for each predicate category given the high-level object category. They are used to generate the corresponding attention maps on the right.}
  \label{fig:vis}
\vspace{-0.5em}
\end{figure*}

\section{Related Work}
\textbf{Visual Relation Detection (VRD)} aims to predict the relationships of given subject-object pairs, which can be viewed as a pair-wise classification task and have been widely studied in the image domain, \eg, scene graph generation (SGG)~\cite{xu2017scene,tang2020unbiased,li2022devil,teng2022structured} and human-object interaction (HOI) detection~\cite{kato2018compositional,kim2020detecting,liao2022gen}. Previous solutions mainly focus on learning representations from the training samples on pre-defined categories, which may suffer from noisy annotations~\cite{li2022devil} or long-tailed predicate distribution~\cite{tang2020unbiased,chen2023addressing} and are far from the needs of the real-world scenarios. Recently, some attempts~\cite{hetao2022,gao2023compositional} adopted prompt-tuning~\cite{liu2023pre} to predict unseen categories during inference. However, since the learnable prompts may be overfitting when trained on seen categories, their performance is sensitive to the split of seen/unseen categories~\cite{gao2022open}. In contrast, our method can predict the relationships directly without any training samples, and has better interpretability and generalization ability, especially in rare informative relation categories.

\textbf{Zero-shot Visual Recognition} enables the model to recognize new categories that it has never seen during training, which is one of the research hotspots in the vision community. Aligning visual representations to pre-trained word embeddings (\eg, Word2Vec~\cite{mikolov2013efficient} and GloVe~\cite{pennington2014glove}) is an intuitive and feasible way to achieve this goal~\cite{wang2018zero}. More recently, VLMs, which use contrastive learning~\cite{chuang2020debiased} to learn a joint space for vision and language, have demonstrated their impressive zero-shot ability~\cite{radford2021learning}. Therefore, many zero-shot works~\cite{zareian2021open,gu2022open,wang2023learning} adopted such VLMs as their basic component to use the knowledge of the learned joint space. However, most of them only utilized the \emph{class name} of unseen categories during inference, which makes an over-strong assumption that the text encoder project proper embeddings with only category names~\cite{menon2022visual}. Then, \citet{menon2022visual} proposed to query LLMs for the rich context of additional information. Nonetheless, it is non-trivial to apply such paradigms to VRD as discussed in Sec.~\ref{sec:intro}. To the best of our knowledge, we are the first to leverage both LLMs and VLMs for VRD in an efficient, effective, and explainable way.

\section{Conclusion}
In this paper, we proposed a novel approach for zero-shot Visual Relationship Detection (VRD) that leverages large language models (LLMs) to generate detailed and informative descriptions of visual cues for each relation category. The proposed method addresses the limitations of traditional class-based prompts and enhances the discriminability of similar relation categories by incorporating specific visual cues. Moreover, we introduced a chain-of-thought method that breaks down the problem into smaller, more manageable pieces, allowing the LLM to generate a series of rationales for each visual cue and ultimately leading to reasonable weights. Our experiments on four benchmark datasets demonstrated the effectiveness and interpretability of our method.
% \subsection{Societal Impact}
% \subsection{Limitations}

\noindent\textbf{Acknowledgement.} This work was supported by the National Key Research \& Development Project of China (2021ZD0110700), the National Natural Science Foundation of China (U19B2043, 61976185),  and the Fundamental Research Funds for the Central Universities (226-2023-00048). Long Chen is supported by HKUST Special Support for Young Faculty (F0927), and HKUST Sports Science and Technology Research Grant (SSTRG24EG04).

\bibliographystyle{unsrtnat}
\bibliography{ref}

\newpage
\renewcommand{\thesection}{\Alph{section}}
\vspace{-10em}
\section*{Appendix}
This supplementary document is organized as follows:
\pagestyle{empty}
\thispagestyle{empty} 
\begin{itemize}
\item The details about stimulated spatial images generation mentioned in Sec.~\textcolor{blue}{2.1} is shown in Sec.~\ref{sec:a}.
\item The prompts for the high-level object category generation (\cf, Sec.~\textcolor{blue}{2.2.1}), visual cue description (\cf, Sec.~\textcolor{blue}{2.2.1}), visual weight determination (\cf, Sec.~\textcolor{blue}{2.2.2}), and filtering strategy (\cf, Sec.~\textcolor{blue}{3.1}) are presented in Sec.~\ref{sec:b}.
\item The implementation details mentioned in Sec.~\textcolor{blue}{3.1} are provided in Sec.~\ref{sec:c}.
\item Additional experimental and qualitative results are reported in Sec.~\ref{sec:d}.
\item The broader impacts of the proposed method are discussed in Sec.~\ref{sec:e}.
\item The limitations of the proposed method are presented in Sec. \ref{sec:f}.
\end{itemize}

\setcounter{table}{4}
\setcounter{figure}{6}

\section{Stimulated Spatial Images Generation}
\label{sec:a}
We propose to simulate the spatial relationship between the subject and object by generating a finite set of spatial images, as mentioned in Sec.~\textcolor{blue}{2.1}. Each spatial image represents the bboxes of the subject and object, where the subject's bounding box is visually denoted by a red box, and the object's bounding box is denoted by a green box. We define four essential attributes, namely shape, size, relative position, and distance, to describe the spatial relationships between the subject and object. These attributes are calculated based on various characteristics, including the aspect ratio $\rho$ and area $A$ of the bounding boxes, the cosine similarity $sim(\cdot,\cdot)$, and the Euclidean distance $d$ between their centers. By assigning different values to these attributes, we can generate a diverse set of simulated spatial images. Given a specific triplet, we calculate the value of each attribute based on the characteristics of the subject and object. Next, we search for the most suitable spatial image in the simulated set by matching these attribute values. This matching process involves comparing the calculated attributes of the triplet with the corresponding attribute ranges in the simulated set. For instance, the aspect ratios and areas of the subject and object bounding boxes determine their shape and size attributes, while the cosine similarity and Euclidean distance between their centers contribute to the relative position and distance attributes. By utilizing this approach, we can effectively simulate various spatial relationships between the subject and object to improve computing efficiency. The detailed procedures of the algorithm are provided in Algorithm~\ref{alg:spatial_simulation}.
    
\begin{algorithm}[H]
\caption{Spatial Relationship Simulation}
\label{alg:spatial_simulation}
\begin{algorithmic}[1]
    \REQUIRE Bounding boxes $b_s$ and $b_o$ for the subject and object of the triplet.
    \ENSURE Spatial image $I_t$ corresponding to the triplet.
    
    \STATE \textbf{Step 1: Generate simulated spatial image set}
    \STATE Define the attributes: shape, size, relative position, and distance.
    \STATE Specify the corresponding value intervals for each attribute.
    \begin{itemize}
    \item Shape: horizontal, vertical, and square denoted as $\{\mathcal{H}, \mathcal{V},\mathcal{Q}\}$.
    \item Size: small, medium, and large denoted as $\{\mathcal{S}, \mathcal{M},\mathcal{L}\}$.
    \item Relative position:  above ($\uparrow$), below ($\downarrow$), left ($\leftarrow$), right ($\rightarrow$), top-left ($\nwarrow$), top-right ($\nearrow$), bottom-left ($\swarrow$), and bottom-right ($\searrow$).
    \item Distance: small, medium, large denoted as $\{\mathcal{S}, \mathcal{M},\mathcal{L}\}$.
    \end{itemize}
    \STATE Draw spatial images by combining all attribute values to create a set of simulated spatial images denoted as $\{I_i\}$. 
    
    \STATE \textbf{Step 2: Calculate attributes of the given triplet}
    \STATE Calculate the centers $c_s$ and $c_o$ of the subject and object bounding boxes, respectively. 
    \STATE Calculate the aspect ratios $\rho_s = \frac{{\text{width}(b_s)}}{{\text{height}(b_s)}}$ and $\rho_o = \frac{{\text{width}(b_o)}}{{\text{height}(b_o)}}$ of the subject and object bounding boxes, respectively. 
    \STATE Calculate the areas $A_s = \text{width}(b_s) \times \text{height}(b_s)$ and $A_o = \text{width}(b_o) \times \text{height}(b_o)$ of the subject and object bounding boxes, respectively. 
    \STATE Calculate the cosine similarity $sim(c_s,c_o) = \frac{{c_s \cdot c_o}}{{\|c_s\| \cdot \|c_o\|}}$ and the Euclidean distance $d_{s,o} = \sqrt{(c_s - c_o)^2}$ between the centers $c_s$ and $c_o$.
    
    \STATE \textbf{Step 3: Match a spatial image in the simulated set}
    \STATE Find the shape and size intervals $S_s$, $S_o$, $L_s$, and $L_o$ that $\rho_s$, $\rho_o$, $A_s$, and $A_o$ belong to. 
    \STATE Find the relative position and distance interval that $\text{sim}(c_s,c_o)$ and $d_{s,o}$ belong to. 
    \STATE Match the spatial image $I_t$ in $\{I_i\}$ by combining the appropriate attributes. 
    \RETURN $I_t$
\end{algorithmic}
\end{algorithm}

\section{Prompts}
\label{sec:b}
In this section, we present prompts for high-level object category generation (\cf, Sec.~\textcolor{blue}{2.2.1}), visual cue description (\cf, Sec.~\textcolor{blue}{2.2.1}), visual weight determination (\cf, Sec.~\textcolor{blue}{2.2.2}), and unreasonable predicate filtering (\cf, Sec.~\textcolor{blue}{3.1}).

\noindent\textbf{High-level Object Class Generation Prompt.} To facilitate the classification of low-level object categories into high-level object categories, we provide the following prompt:
\begin{Verbatim}[commandchars=\\\{\}]
\textcolor{orange}{Given the low-level object categories: [ALL OBJ CLS].}
\textcolor{orange}{Please classify each low-level object category into high-level object}
\textcolor{orange}{categories ["human", "animal", "product"] based on their most common}
\textcolor{orange}{semantics in visual relation detection.} 
\textcolor{orange}{Ensure that body parts and similar categories are not classified as "human". }
\textcolor{orange}{Note that human beings engaged in certain activities must be classified as}
\textcolor{orange}{"human"!}
\end{Verbatim}
In this prompt, we provide a list of low-level object categories: [ALL OBJ CLS] that need to be categorized into high-level object categories(\ie, ``human'', ``animal'', and ``product''). The prompt instructs the LLMs to assign the low-level object categories to the most appropriate high-level object category based on their prevalent semantics in visual relation detection. This prompt guides the model to understand the distinctive characteristics and visual cues associated with different object categories, contributing to accurate descriptions of visual cue for each relation category.

\noindent\textbf{Visual Cue Description Prompt.} We present guided relation component description prompt for generating the descriptions of visual cue, specifically designed for the relation class ``REL CLS'' when provided with the High-Level (HL) categories of the subject and object, \ie, ``SUB HL CLS'' and ``OBJ HL CLS''. The prompt is structured as follows:
\begin{Verbatim}[commandchars=\\\{\}]
\textcolor{orange}{
Known: a visual triplet is formulated as [subject, predicate, object].}
\textcolor{orange}{
Note that:}
  \textcolor{orange}{[position] must not include nouns other than subject and object!}
  \textcolor{orange}{[position] must contain}
  \textcolor{orange}{
  [orientation: ("above", "below", "left", "right", "inside"),}
    \textcolor{orange}{shape: ("horizontal", "vertical", "square"),}
    \textcolor{orange}{distance: ("small distance", "mid distance", "large distance")]!}
    
\textcolor{orange}{Describe the visual features of the predicate "sitting on" in a photo, }
\textcolor{orange}{when subject belongs to [human], object belongs to [product]:} 
\textcolor{orange}{  [subject]:}
\textcolor{orange}{    - with legs.}
\textcolor{orange}{    - with hip.}
\textcolor{orange}{  [object]:}
\textcolor{orange}{    - with flat surface.}
\textcolor{orange}{  [position]:}
\textcolor{orange}{    - square subject above horizontal object with a small distance.}

\textcolor{orange}{Describe the visual features of the predicate "REL CLS" in a photo, }
\textcolor{orange}{when subject belongs to [SUB HL CLS], object belongs to [OBJ HL CLS]:}
\end{Verbatim}

The prompt is divided into four distinct parts: setting, constraint, example, and question.
\textbf{Setting:} The setting (\ie, ``\texttt{\textcolor{orange}{Known...}}'') provides specific roles and known conditions for the LLMs to operate within.
\textbf{Constraint:} The constraint (\ie, ``\texttt{\textcolor{orange}{Note that...}}'') outlines some limitations or constraints on the output generated by the LLMs.
\textbf{Example:} The example (\ie, the example of ``sitting on'') serves as a guide for the model to produce similar output in an in-context learning~\citep{brown2020language,liu2021makes} manner, which is also generated by LLM.
\textbf{Question:} Finally, the question (\ie, ``\texttt{\textcolor{orange}{Describe...}}'') prompts the model to generate a description of the visual features that are specific to the relation being considered.
This comprehensive prompt structure aids in more reasonable and standardized generation of visual cue descriptions for subject, object, and spatial components for each relation category.

\noindent\textbf{Visual Cue Weight Prompt.}
The prompt is designed to determine the visual cue weights for subject (SUB CUES), object (OBJ CUES), and spatial (POS CUES) in relation classification. It is structured as follows:
\begin{Verbatim}[commandchars=\\\{\}]
\textcolor{orange}{Suppose you are a relation classification model.}

\textcolor{orange}{Given: subject belongs to [human] and object belongs to [product].}
\textcolor{orange}{The visual features of subject:}
  \textcolor{orange}{["with eyes directed towards the object", "with head upright"].}
\textcolor{orange}{The visual features of object:}
  \textcolor{orange}{
["with visible features such as front, display, or screen"].} 
\textcolor{orange}{The visual features of position:}
  \textcolor{orange}{["subject positioned either above, below, left or right of the object at}
  \textcolor{orange}{a mid distance"].}
\textcolor{orange}{Q: How do you weight these visual features (subject, object, position) to}
\textcolor{orange}{determine the predicate is "looking at"? The sum of weights must be 1.0!
}
\textcolor{orange}{A: Let's think step by step!}
\textcolor{orange}{First, we need to consider the importance of the subject's visual features.} 
\textcolor{orange}{Since the direction of the eyes and head position strongly indicate the }
\textcolor{orange}{focus of attention, we will give them a weight of 0.6. Next, we need to}
\textcolor{orange}{consider the importance of the object's visual features. Since the visible}
\textcolor{orange}{features such as front, display, or screen indicate that the object is} 
\textcolor{orange}{something that can be looked at, we will give them a weight of 0.3. Finally,}
\textcolor{orange}{we need to consider the importance of the position visual features. Since}
\textcolor{orange}{the relative position of the subject and object at a mid-distance helps us}
\textcolor{orange}{understand that the subjects are looking at the object in question, we will}
\textcolor{orange}{give them a weight of 0.1.}
\textcolor{orange}{
Therefore, we can weight these visual features as follows:}
\textcolor{orange}{
Weight("looking at") = 0.6 * Weight(visual features of subject) }
                     \textcolor{orange}{+ 0.3 * Weight(visual features of object)}
                     \textcolor{orange}{+ 0.1 * Weight(visual features of position).}

\textcolor{orange}{Given: subject belongs to [SUB HL CLS] and object belongs to [OBJ HL CLS].}
\textcolor{orange}{The visual features of subject:}
  \textcolor{orange}{[SUB CUES].}
\textcolor{orange}{The visual features of object:}
  \textcolor{orange}{
[OBJ CUES].} 
\textcolor{orange}{The visual features of position:}
  \textcolor{orange}{[POS CUES].}
\textcolor{orange}{Q: How do you weight these visual features (subject, object, position) to}
\textcolor{orange}{determine the predicate is "REL CLS"? The sum of weights must be 1.0!
}
\textcolor{orange}{A: Let's think step by step!}
\end{Verbatim}
The prompt is also divided into four distinct parts: setting, constraint, example, question.
\textbf{Setting:} The setting (\ie, ``\texttt{\textcolor{orange}{Suppose...}}'') establishes the role and perspective of the model in the task.
\textbf{Constraint:} The constraint (\ie, ``\texttt{\textcolor{orange}{The sum of weights must be 1.0!}}'') provides some limitations or constraints on the output generated by the LLMs.
\textbf{Example:} The example (\ie, the example of determining the predicate ``looking at'') serves as a guide for the LLMs to understand the context and expected output.
\textbf{Question:} The question (\ie, ``\texttt{\textcolor{orange}{How do...}}'') prompts the model to determine the weights assigned to visual cues in order to classify the given predicate.
Additionally, the stepwise prompt ``\texttt{\textcolor{orange}{Let’s think step by step!}}'' guides the LLMs to incrementally analyze the problem and generate rationales, which lead to more reasonable determination of visual cue weights. 

\noindent\textbf{Filter Prompt.} The prompt is used to filter unreasonable \texttt{sub-pred} and \texttt{obj-pred} categories. The prompt for \texttt{sub-pred} is as follows:

\begin{Verbatim}[commandchars=\\\{\}]
\textcolor{orange}{Q: Can the window be sitting on something?}
\textcolor{orange}{After thinking about it, just answer "Yes" or "No"!}
\textcolor{orange}{A: Let's think step by step!} 
\textcolor{orange}{It is possible for a window to be sitting in something, such as a frame}
\textcolor{orange}{or sill.}
\textcolor{orange}{Answer is Yes.}

\textcolor{orange}{Q: Can the \{SUB CLS\} be \{REL CLS\} something?}
\textcolor{orange}{After thinking about it, just answer "Yes" or "No"!}
\textcolor{orange}{A: Let's think step by step!}
\end{Verbatim}
The prompt also adopts an in-context learning approach, leveraging illustrative examples to enhance comprehension of the situation. The stepwise prompt stimulates logical reasoning, facilitating the LLMs in rendering more robust judgments by leveraging the provided information.

\section{Implementation Details}
\label{sec:c}
Our RECODE does not require a training process and can be directly tested on a NVIDIA 2080 Ti GPU. We pre-computed the visual features encoded by CLIP for each bounding box, enabling us to set the batch size to 512. For the LLM, we utilized the GPT-3.5-turbo, a highly performant variant of the GPT model. As for CLIP, we leveraged the OpenAI's publicly accessible resources, specifically opting for the Vision Transformer with a base configuration (ViT-B/32) as the default backbone.

\section{Further Analysis}
\label{sec:d}

\subsection{Comparison with Training-based Methods}

\addtolength{\tabcolsep}{-1.5pt}
\begin{table}[!t]
  \centering
  \caption{Comparison with SOTA VRD methods on the VG dataset. Note that none of these methods can be applied in the \textbf{training-free} zero-shot setting.}
      \vspace{-0.5em}
  \setlength\tabcolsep{0.5pt}
  \resizebox{0.8\textwidth}{!}{
    \begin{tabular}{lcccccc}
    \specialrule{0.1em}{0pt}{0pt}
    \hline
    \multirow{2}[4]{*}{\small{Model}}  & \multirow{2}[4]{*}[0.8ex]{\parbox{1.5cm}{\centering \small{No}\\ \small{Training}}}  & \multirow{2}[4]{*}[1.0ex]{\parbox{1.5cm}{\centering \small{Unseen}\\ \small{Relation}}} & \multirow{2}[4]{*}[1.0ex]{\parbox{2cm}{\centering \small{Training}\\ \small{Data Source}}} & \multicolumn{3}{c}{\small{Predicate Classification}} \\    \cline{5-7}   &   &       &       &      \footnotesize{zR@20} & \footnotesize{zR@50} & \footnotesize{zR@100} \\
    \hline
    \textcolor{gray}{\small{Motifs}}~\cite{zellers2018neural} & \textcolor{gray}{\XSolidBrush}     & \textcolor{gray}{\XSolidBrush}  & \textcolor{gray}{\small{VG}}  & \textcolor{gray}{8.9}   & \textcolor{gray}{15.2}  & \textcolor{gray}{18.5} \\
    \textcolor{gray}{\small{COACHER}}~\cite{kan2021zero} & \textcolor{gray}{\XSolidBrush}      & \textcolor{gray}{\XSolidBrush}   & \textcolor{gray}{\small{VG\& ConceptNet}}    & \textcolor{gray}{28.2}  & \textcolor{gray}{34.1}  & \textcolor{gray}{37.2} \\
    \textcolor{gray}{\small{DPL}}~\cite{li2023decomposed}   & \textcolor{gray}{\XSolidBrush}     & \textcolor{gray}{\XSolidBrush}       & \textcolor{gray}{\small{VG}}  & \textcolor{gray}{6.0}   & \textcolor{gray}{7.7}   & \textcolor{gray}{9.3} \\
    \textcolor{gray}{\small{CaCao}}~\cite{yu2023visually} & \textcolor{gray}{\XSolidBrush}   & \textcolor{gray}{\Checkmark} & \textcolor{gray}{\small{VG\&CC3M\&COCO}}  & \textcolor{gray}{17.2}  & \textcolor{gray}{21.3}  & \textcolor{gray}{23.1} \\
    \hline
    \small{RECODE} & \textcolor{red}{\Checkmark} &  \textcolor{red} {\Checkmark} & -  & 8.2   & 16.1  & 23.2 \\
    \specialrule{0.1em}{0pt}{0pt}
    \hline
    \end{tabular}%
    }
  \label{tab:sota_sgg}%
\end{table}%
 \addtolength{\tabcolsep}{1.5pt}

In this section, we compared the proposed \textbf{training-free} RECODE framework with those well-designed training-based ones in Table~\ref{tab:sota_sgg}. Note that such comparisons are unfair as training-based frameworks can learn the underline patterns and data distribution from the training set. For completeness, we still reported the results and investigate the performance gap between training-based frameworks and RECODE. Specifically, we compared the proposed RECODE with several relevant baselines, including triplet-level zero-shot VRD~\citep{zellers2018neural,kan2021zero}, few-shot VRD~\citep{li2023decomposed}, and category-level zero-shot VRD~\citep{yu2023visually}. Since all of them can not detect relations without training, we reported Zero-shot Recall@K (\textbf{zR@K}), which only calculates the Recall@K for those unseen \textbf{triplet} categories. 
\begin{itemize}
    \item Triplet-level zero-shot VRD methods. Motifs~\citep{zellers2018neural} is a traditional strong baseline without explicitly modeling the nature of zero-shot. COACHER~\citep{kan2021zero} explicitly models the nature of zero-shot, and takes the power of the common sense from ConceptNet resulting in better performance.
    \item Few-shot VRD methods. DPL~\citep{li2023decomposed} is a few-shot baseline, which mainly investigates making predictions with a few examples (here we evaluate 1-shot).
    \item Category-level zero-shot VRD methods. CaCao~\citep{yu2023visually} also explicitly models the nature of zero-shot, and leverages language information from captions of CC3M and COCO for enhanced performance.
\end{itemize}
Surprisingly, even without training, RECODE still achieves competitive results, with zR@20, zR@50, and zR@100 of 8.2\%, 16.1\%, and 23.2\%, respectively. This signifies its potential in handling unseen categories, due to the effective visual cues and inference mechanisms. 

\subsection{Ablation on Different Class-based Prompts}

\addtolength{\tabcolsep}{-1.5pt}
\begin{table}[htbp]
  \centering
  \caption{Ablation studies of different class-based prompts on the test set of VG}
    \renewcommand\arraystretch{1.1}
  \setlength\tabcolsep{2pt}
    \begin{tabular}{l|l|ccc|ccc}
    \specialrule{0.1em}{0pt}{0pt}
    \hline
    \multicolumn{1}{l|}{\multirow{2}[4]{*}{Class-based Prompt}} & \multicolumn{1}{c|}{\multirow{2}[4]{*}{Method}} & \multicolumn{6}{c}{Predicate Classification} \\
\cline{3-8}    \multicolumn{1}{c|}{} &       & \small{R@20} & \small{R@50} & \small{R@100} & \small{mR@20} & \small{mR@50} & \small{mR@100} \\
    \hline
    \multirow{2}[2]{*}{[REL CLS]-ing/ed} & CLS$^\star$   & 7.5   & 13.7  & 19.4  & 9.1   & 15.9  & 24.0 \\
          & RECODE$^\star$ & \textbf{10.6} & \textbf{18.3} & \textbf{25.0} & \textbf{10.7} & \textbf{18.7} & \textbf{27.8} \\
    \hline
    \multirow{2}[2]{*}{a photo of [REL CLS]} & CLS$^\star$   & 11.7  & 19.2  & 26.2  & 10.9  & 19.4  & 27.1 \\
          & RECODE$^\star$ & \textbf{13.5} & \textbf{21.8} & \textbf{28.8} & \textbf{12.2} & \textbf{19.6} & \textbf{28.4} \\
    \specialrule{0.1em}{0pt}{0pt}
    \hline
    \end{tabular}%
  \label{tab:cls_prompts}%
\end{table}%
 \addtolength{\tabcolsep}{1.5pt}

We conducted an ablation study to investigate the impact of different class-based prompts on zero-shot VRD performance. The class-based prompts were manually designed to generate text embedding for relation classification. We compared two types of class-based prompts: 1) ``[REL-CLS]-ing/ed'' prompt, where [REL-CLS] represents the name of relation category. For example, for the relation class ``milk'', the prompt would be ``milking''. 2) ``a photo of [REL-CLS]'' prompt. For example, for the relation class ``riding'', the prompt would be ``a photo of riding''.

Table \ref{tab:cls_prompts} summarized the results. Our method achieved improved scores across various metrics for both types of prompts. With the ``[REL-CLS]-ing/ed'' prompt, we observed significant gains (3.1\% to 5.6\%) on R@K. Similarly, when using the "a photo of [REL-CLS]" prompt, we achieved the highest R@100 score of 28.8\% and mR@100 of 28.4\%. These results indicated that our method consistently outperforms the CLS baseline, regardless of the specific prompt type used. The effectiveness of our method suggested a promising solution for zero-shot VRD tasks.

\subsection{Interpretability Analysis}
\begin{figure*}[!t]
  \centering
    \includegraphics[width=1\linewidth]{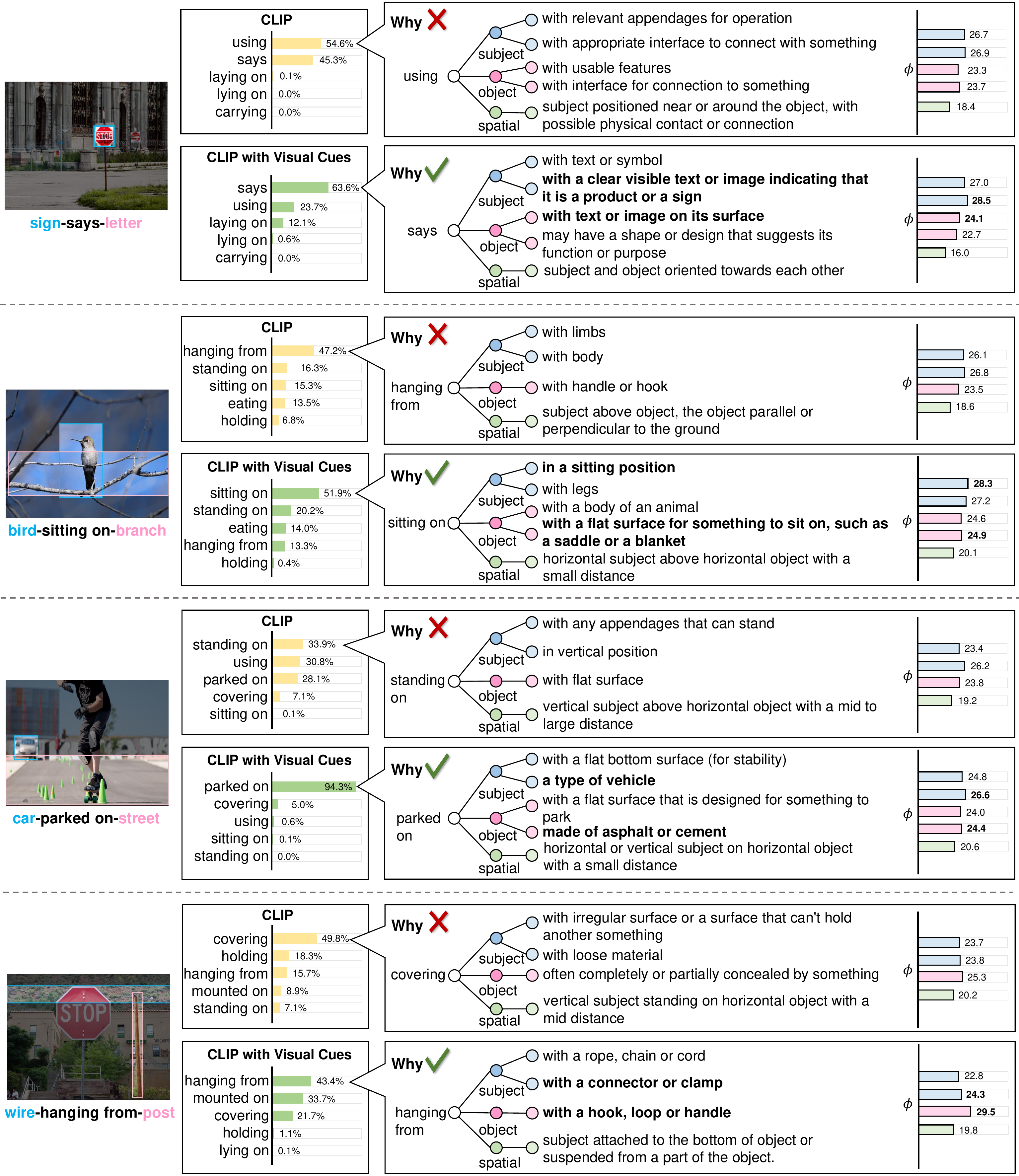}
    \caption{A comparative analysis of predictions made by RECODE and a baseline with class-based prompts on the test set of VG. It illustrates how our method offers interpretability to the VRD through the similarity $\phi$ between the image and the description-based prompts.}
    \label{fig:more_example}
\end{figure*}

To gain a deeper understanding of the interpretability of our RECODE in VRD, we conducted an in-depth analysis comparing its predictions with CLS baseline that utilizes only class-based prompts. By evaluating the similarity between the description-based prompts and the corresponding visual features, we revealed the underlying reasons for the accuracy of RECODE's predictions and the inaccuracies of the CLIP baseline.

Figure \ref{fig:more_example} presented qualitative comparisons of RECODE and the CLS baseline on challenging examples from the VG dataset. Our description-based prompts significantly improve CLIP's understanding of the various relation categories, leading to more accurate predictions. Taking the top image of Figure \ref{fig:more_example} as an example, RECODE accurately predicts the ``says'' relation category by identifying the presence of visual features associated with ``text or image''. In contrast, the failure case of the ``using'' relationship category predicted by the CLS baseline can be attributed to the absence of distinctive visual features related to ``usable feature'', as highlighted by our description-based prompts.

\section{Broader Impacts}
\label{sec:e}
Like every coin has two sides, using our method will have both positive and negative impacts. 

\noindent\textbf{Positive Impacts.} \textbf{Firstly}, RECODE emphasizes the importance of pairwise recognition, encouraging researchers to develop more diverse and comprehensive recognition models. By focusing on the relationships between object pairs, we inspire the exploration of a broader range of relationship types and promote a deeper understanding of complex interactions between objects (\eg, n-tuple interaction), especially in the zero-shot setting without any extra training stage. \textbf{Secondly}, our method introduces the incorporation of spatial information in visual relation detection. By considering spatial cues and relationships between objects, we highlight the significance of spatial information in understanding object interactions. This not only improves the accuracy of relation detection but also encourages researchers to explore the integration of other useful auxiliary information. This can include incorporating contextual information, temporal relationships, or other relevant cues that can enhance recognition performance. \textbf{Thirdly}, our method promotes the use of Chain-of-Thought prompting with LLMs for weight assignment in recognition tasks. By leveraging the knowledge and capabilities of LLMs, we enable the generation of more informed and reasonable weights for different components of the recognition process. This improves the interpretability of the recognition results and opens up new possibilities for utilizing the vast knowledge and capabilities of language models to enhance recognition systems.

\noindent\textbf{Negative Impacts.} However, we also acknowledge that there are potential negative impacts associated with the use of our method. For example, the reliance on LLMs could lead to the perpetuation of biases and inequalities present in the data used to pre-train these models.

In conclusion, the proposed method for zero-shot visual relation detection brings about positive impacts by inspiring more complex recognition models under the zero-shot setting, highlighting the significance of contextual cues, and promoting the use of LLMs for weight assignment. It is essential to continue exploring ways to address potential negative impacts and ensure the responsible and ethical use of these advancements in our community.

\section{Limitations}
\label{sec:f}
As the first zero-shot visual relation detection work using LLMs, our method still has some limitations: \textbf{1) Firstly}, we did not specifically evaluate spatial relation categories (\eg, ``on'', ``under'') and ownership relation categories (\eg, ``belong to''). In this work, our method mainly focuses on classifying semantic predicate groups based on visual cue descriptions. However, by extensive empirical results, these spatial and ownership relationships can be easily predicted from only spatial positions or object categories. \textbf{2) Secondly}, our framework assumes the availability of ground truth bounding boxes and object categories for relation classification. However, in real-world scenarios, object detection can introduce errors or uncertainties. \textbf{3) Thirdly}, to avoid overmuch queries to LLMs, our approach proposes a trade-off solution and only relies on coarse-grained triplet category descriptions. However, this simplification may not capture fine-grained nuances in different visual relationships. Using more detailed and comprehensive descriptions (with more LLM queries) could potentially further improve the performance. \textbf{4) Fourthly}, the accuracy and correctness of the visual cue descriptions are not guaranteed. Despite efforts to ensure quality, errors or incomplete information may be present. It is essential to validate and verify cue descriptions for reliable results.

% \bibliographystyle{unsrtnat}
% \bibliography{ref}

% \bibliographystyle{unsrtnat}
% \bibliography{ref}

% References follow the acknowledgments in the camera-ready paper. Use unnumbered first-level heading for
% the references. Any choice of citation style is acceptable as long as you are
% consistent. It is permissible to reduce the font size to \verb+small+ (9 point)
% when listing the references.
% Note that the Reference section does not count towards the page limit.
% \medskip

% {
% \small

% [1] Alexander, J.A.\ \& Mozer, M.C.\ (1995) Template-based algorithms for
% connectionist rule extraction. In G.\ Tesauro, D.S.\ Touretzky and T.K.\ Leen
% (eds.), {\it Advances in Neural Information Processing Systems 7},
% pp.\ 609--616. Cambridge, MA: MIT Press.

% [2] Bower, J.M.\ \& Beeman, D.\ (1995) {\it The Book of GENESIS: Exploring
%   Realistic Neural Models with the GEneral NEural SImulation System.}  New York:
% TELOS/Springer--Verlag.

% [3] Hasselmo, M.E., Schnell, E.\ \& Barkai, E.\ (1995) Dynamics of learning and
% recall at excitatory recurrent synapses and cholinergic modulation in rat
% hippocampal region CA3. {\it Journal of Neuroscience} {\bf 15}(7):5249-5262.
% }

%%%%%%%%%%%%%%%%%%%%%%%%%%%%%%%%%%%%%%%%%%%%%%%%%%%%%%%%%%%%

% \end{document}
%%%%%%%%%%%%%%%%%%%%%%%%%%%%%%%%%%%%%%%%%%%%%%%%%%%%%%%%%%%%
\end{document}